\documentclass[times,twocolumn,final,authoryear]{elsarticle}

\usepackage{ycviu}
\usepackage{framed,multirow}

\usepackage{amsmath,amssymb} %
\usepackage{latexsym}
\usepackage[linesnumbered,vlined,ruled]{algorithm2e}

\usepackage{url}
\usepackage{xcolor}
\definecolor{newcolor}{rgb}{.8,.349,.1}

\journal{Computer Vision and Image Understanding}

\begin{document}

\ifpreprint
  \setcounter{page}{1}
\else
  \setcounter{page}{1}
\fi

\begin{frontmatter}

\title{Curriculum self-paced learning for cross-domain object detection}

\author[1,3]{Petru \snm{Soviany}} 
\author[1,2,4]{Radu Tudor \snm{Ionescu}\corref{cor1}}
\cortext[cor1]{Corresponding author: }
\ead{raducu.ionescu@gmail.com}

\author[3]{Paolo \snm{Rota}}
\author[3]{Nicu \snm{Sebe}}

\address[1]{Department of Computer Science, University of Bucharest, 14 Academiei Street, Bucharest 010014, Romania}
\address[2]{Romanian Young Academy, University of Bucharest, 90 Panduri Street, Bucharest 050663, Romania}
\address[3]{Department of Information Engineering and Computer Science, University of Trento, 9 Sommarive Street, Povo-Trento 38123, Italy}
\address[4]{SecurifAI, 21 Mircea Voda, Bucharest 030662, Romania}

\received{1 May 2013}
\finalform{10 May 2013}
\accepted{13 May 2013}
\availableonline{15 May 2013}
\communicated{S. Sarkar}

\begin{abstract}
Training (source) domain bias affects state-of-the-art object detectors, such as Faster R-CNN, when applied to new (target) domains. To alleviate this problem, researchers proposed various domain adaptation methods to improve object detection results in the cross-domain setting, e.g.~by translating images with ground-truth labels from the source domain to the target domain using Cycle-GAN. On top of combining Cycle-GAN transformations and self-paced learning in a smart and efficient way, in this paper, we propose a novel self-paced algorithm that learns from easy to hard. 
Our method is simple and effective, without any overhead during inference. It uses only pseudo-labels for samples taken from the target domain, i.e. the domain adaptation is unsupervised. We conduct experiments on four cross-domain benchmarks, showing better results than the state of the art. We also perform an ablation study demonstrating the utility of each component in our framework. Additionally, we study the applicability of our framework to other object detectors. Furthermore, we compare our difficulty measure with other measures from the related literature, proving that it yields superior results and that it correlates well with the performance metric.
\end{abstract}


\end{frontmatter}



\section{Introduction}

Machine learning models exhibit poor performance when the test (target) data is sampled from a different domain than the training (source) data, mainly due to the distribution gap (domain shift) between  different domains. Domain shift is a well-studied problem in the broad area of machine learning~\citep{Chang-AAAI-2017,Chen-CVPR-2018,Fernando-ICCV-2013,Ganin-JMLR-2016,Khodabandeh-ICCV-2019,Saito-CVPR-2019,Sener-NIPS-2016,Sun-AAAI-2016,Zheng-CVPR-2020,Zhuang-IJCAI-2013}, attracting a lot of attention in computer vision~\citep{Chang-AAAI-2017,Fernando-ICCV-2013,Rozantsev-TPAMI-2018,Saito-CVPR-2019,Sener-NIPS-2016,Shu-CIKM-2015,Sun-AAAI-2016,Tzeng-CVPR-2017} and related fields~\citep{Daume-ACL-2007,Fernandez-JAIR-2016,Ionescu-ICONIP-2018,Pan-WWW-2010,Zhuang-IJCAI-2013}. To understand and address the domain gap, which occurs when labeled data in a target domain is scarce or not even available, researchers have studied the behavior of machine learning models in the cross-domain setting~\citep{Franco-KBS-2015,Lui-IJCNLP-2011} and proposed several domain adaptation methods~\citep{Chang-AAAI-2017,Fernandez-JAIR-2016,Ganin-JMLR-2016,Rozantsev-TPAMI-2018,Sener-NIPS-2016,Shu-CIKM-2015,Tzeng-CVPR-2017}.

\begin{figure*}[!t]
\begin{center}
\includegraphics[width=0.78\linewidth]{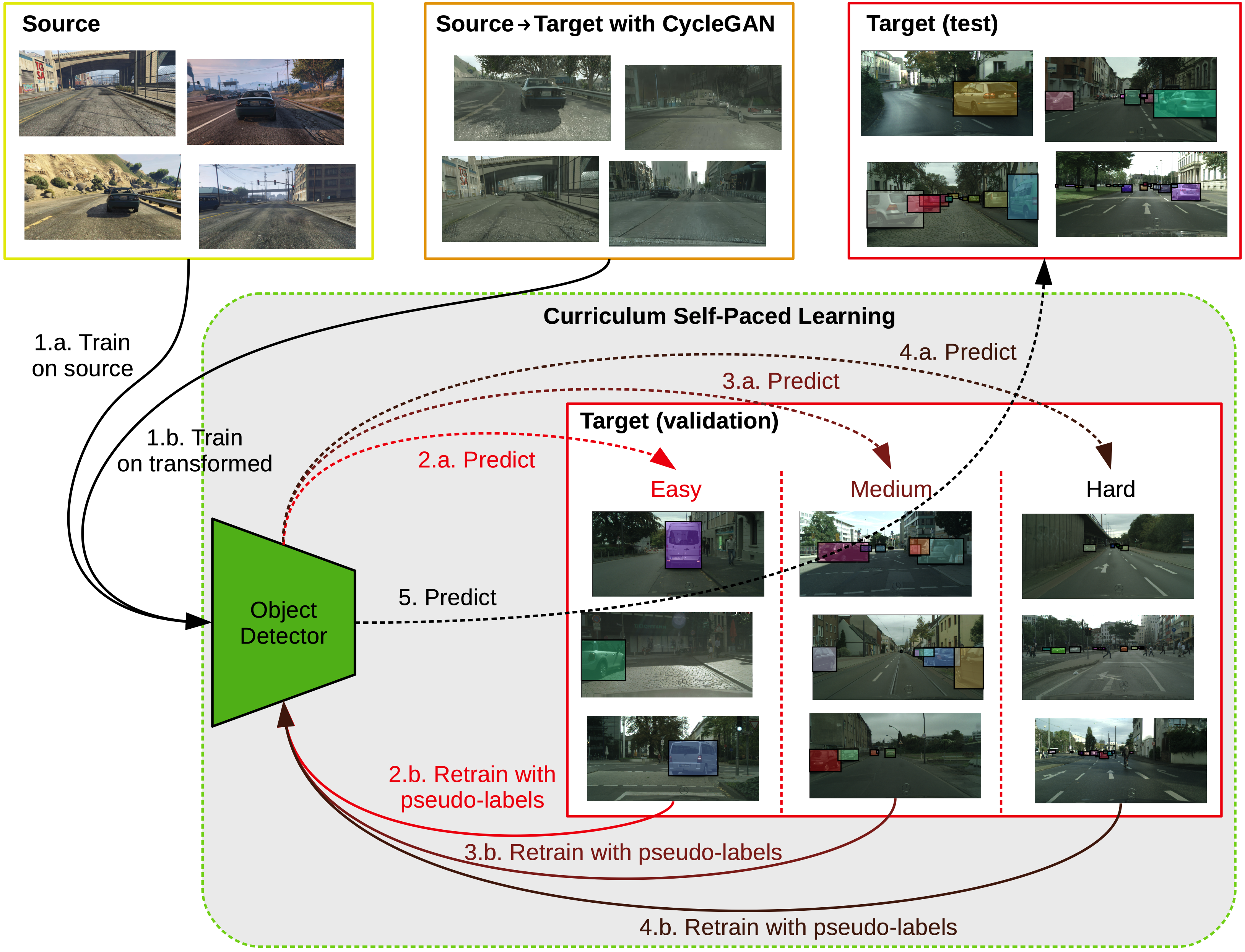}
\end{center}
\vspace*{-0.3cm}
\caption{Our curriculum self-paced learning approach for object detection. In the initial training stage (step 1.a), the object detector is trained on source images with ground-truth labels. In step 1.b, the object detector is further trained on source images translated by Cycle-GAN~\citep{Zhu-ICCV-2017} to resemble images from the target domain. In steps 2, 3 and 4, the object detector is fine-tuned on real target images (different from those included in the test set), using the bounding boxes and the labels predicted by the current detector. In step 5, the model makes its predictions on the target test set for the final evaluation. Best viewed in color.}
\label{fig_pipeline}
\end{figure*}

Domain adaptation methods can be divided into supervised and unsupervised approaches. While supervised approaches use small subsets of labeled samples from the target domain~\citep{Cozma-ACL-2018,Inoue-CVPR-2018}, the unsupervised ones use only unlabeled target samples~\citep{Chen-CVPR-2018,Ganin-JMLR-2016,Guo-ICDM-2012,Raj-BMVC-2015,Saito-CVPR-2019,Shan-NC-2019,Sener-NIPS-2016,Tzeng-CVPR-2017}. In this paper, we propose an unsupervised domain adaptation method for object detection. In cross-domain object detection~\citep{Chen-CVPR-2018,Khodabandeh-ICCV-2019,Raj-BMVC-2015,Saito-CVPR-2019,Shan-NC-2019,Zheng-CVPR-2020,Zhu-CVPR-2019}, an object detector is trained on data from a source domain and tested on data from a target domain, sharing the same object categories. Adapting the object detector for the cross-domain setting can provide the means to train robust models on large-scale data sets, that can be cheaply collected, but are outside the target domain. One such example is training object detectors for real-world street scenes, e.g. Cityscapes~\citep{Cordts-CVPR-2016}, by using artificially generated scenes from realistic video games, e.g. Sim10k~\citep{Johnson-Roberson-ICRA-2017}. We actually test our domain-adapted detector in this setting, which has immediate application in autonomous driving.

We propose a novel curriculum self-paced learning approach in order to adapt an object detector to the target domain. In self-paced learning, the model learns from its own predictions (pseudo-labels) in order to gain additional accuracy. Since we use image samples from the target domain during inference, the model has the opportunity to learn domain-specific features, thus adapting itself to the target domain. However, the main problem in self-paced learning is that the model can be negatively influenced by the noisy pseudo-labels, i.e.~prediction errors. In order to alleviate this problem, we propose an effective combination of two approaches. In order to reduce the labeling noise level, we apply a domain-adaptation approach that relies only on ground-truth labels, before starting the self-paced learning stage. The approach consists in training a Cycle-consistent Generative Adversarial Network (Cycle-GAN)~\citep{Zhu-ICCV-2017} in order to learn how to transform images from the source domain to the target domain. The adaptation consists in fine-tuning the object detector on source images that are translated by Cycle-GAN to look like target images (see Figure~\ref{fig_pipeline} for some translated samples). In the experiments, we show that reducing the labeling noise before self-paced learning is indeed helpful, but still not satisfactory. 
We hypothesize that the labeling noise inherently induced by the prediction errors is proportional to the difficulty of the images. Following this intuition, we perform self-paced learning starting with the easier images, gradually adding more and more difficult image samples, inspired by the curriculum learning paradigm~\citep{Bengio-ICML-2009}, as shown in Figure~\ref{fig_pipeline}. Our hypothesis turns out to be supported by the empirical results, confirming the utility of our curriculum self-paced learning method. In order to estimate the difficulty of each image sample, we employ a score given by the number of detected objects divided by the average area of their bounding boxes. This is inspired by the previous work of~\cite{Ionescu-CVPR-2016}, which found that image difficulty is directly proportional to the number of objects and inversely proportional to the average bounding box area. However, we empirically show that our metric provides superior results compared with the difficulty metric of~\cite{Ionescu-CVPR-2016}.

We evaluate our curriculum self-paced learning approach on four cross-domain benchmarks, Sim10k$\rightarrow$Cityscapes, KITTI$\rightarrow$Cityscapes, PASCAL VOC 2007$\rightarrow$Clipart1k and PASCAL VOC 2007+2012$\rightarrow$Clipart1k, comparing it with recent state-of-the-art methods~\citep{Chen-CVPR-2018,Inoue-CVPR-2018,Khodabandeh-ICCV-2019,Saito-CVPR-2019,Shan-NC-2019,Zheng-CVPR-2020,Zhu-CVPR-2019}, whenever possible. The empirical results indicate that our approach provides the highest absolute gains (with respect to the baseline Faster R-CNN detector), surpassing all considered competitors~\citep{Chen-CVPR-2018,Inoue-CVPR-2018,Khodabandeh-ICCV-2019,Saito-CVPR-2019,Shan-NC-2019,Zheng-CVPR-2020,Zhu-CVPR-2019}. Furthermore, we consider that our performance gains of $+17.01\%$ on Sim10k$\rightarrow$Cityscapes, $+12.34\%$ on KITTI$\rightarrow$Cityscapes, $+8.91\%$ on PASCAL VOC 2007$\rightarrow$Clipart1k and $+11.69\%$ on PASCAL VOC 2007+2012$\rightarrow$Clipart1k are significant. Our experiments also include ablation results, showing how various components and parameters influence our performance level.

The rest of this paper is organized as follows. In Section~\ref{sec_related}, we present the related works on domain adaptation, self-paced learning, curriculum learning and cross-domain object detection. Our curriculum self-paced learning method is detailed in Section~\ref{sec_method}. The comparative and ablation experiments are presented in Section~\ref{sec_experiments}. Finally, we draw our conclusion and discuss future work in Section~\ref{sec_conclusion}.

\section{Related Work}
\label{sec_related}

\noindent
{\bf Domain Adaptation.}
Domain adaptation is the task of fitting a model trained on a source distribution to a different target distribution. One immediate use case is the elimination of the costly human labeling process by automatically generating artificial training data, e.g.~object detectors for autonomous driving could be trained on video game scenes. Domain adaptation has been extensively studied in cross-domain classification problems. The corresponding methods can be roughly categorized into cross-domain kernels~\citep{Duan-TPAMI-2011,Ionescu-ICONIP-2018}, sub-space alignment~\citep{Fernando-ICCV-2013}, second-order statistics alignment~\citep{Sun-AAAI-2016}, adversarial adaptation~\citep{Ganin-JMLR-2016,Tzeng-CVPR-2017}, graph-based methods~\citep{Chang-AAAI-2017,Arun-SDM-2017,Pan-WWW-2010,Ponomareva-RANLP-2013}, probabilistic models~\citep{Luo-EMNLP-2015,Zhuang-IJCAI-2013}, knowledge-based models~\citep{Bollegala-KDE-2013,Franco-KBS-2015} and joint optimization frameworks~\citep{Long-KDE-2014}. To our knowledge, curriculum domain adaptation has not been extensively studied in literature~\citep{Zhang-ICCV-2017}.~\cite{Zhang-ICCV-2017} proposed a curriculum domain adaptation method for semantic segmentation. They applied curriculum over tasks, starting with the easier ones, which are less sensitive to the domain gap than semantic segmentation. Different from~\cite{Zhang-ICCV-2017}, we assign a difficulty score to each image sample, thus applying curriculum over samples. Furthermore, we employ Cycle-GAN as a way to reduce the labeling noise before our curriculum self-paced learning stage.


\noindent
{\bf Curriculum Learning.}
~\cite{Bengio-ICML-2009} introduced easy-to-hard strategies to  train machine learning models, showing that the standard learning paradigm used in human educational systems also applies to artificial intelligence. Curriculum learning represents the general class of algorithms in which the training data are fed gradually, from easy-to-difficult, taking into consideration some difficulty measure. Curriculum learning has been successfully applied to different tasks, including semi-supervised image classification~\citep{Gong-TIP-2016}, language modeling~\citep{Graves-ICML-2017}, weakly-supervised object detection~\citep{Wang-ICPR-2018,Zhang-IJCV-2019}, weakly supervised object localization~\citep{Ionescu-CVPR-2016,Li-BMVC-2017}, person re-identification~\citep{Wang-ECCV-2018} and image generation~\citep{Doan-AAAI-2019,Soviany-WACV-2020}. To the best of our knowledge, curriculum learning has not been applied to cross-domain object detection. In our work, we apply curriculum over target instances that are annotated with pseudo-labels given by the object detector at hand, resulting in a method that combines curriculum and self-paced learning.

\noindent
{\bf Self-Paced Learning.}
In self-paced learning, machine learning models learn from their own labels while taking into consideration the predictions with high confidence first. Self-paced learning is similar to curriculum learning because the training samples are presented in a meaningful order. \cite{Kumar-NIPS-2010} argued that their self-paced learning approach differs from curriculum learning, as it does not rely on an external difficulty measure, but on simultaneously selecting easy samples and on updating the parameters in an iterative manner, based on the actual performance. \cite{Jiang-AAAI-2015} introduced self-paced curriculum learning as an optimization problem taking into account both prior knowledge and knowledge gained during the learning process. We propose a similar approach for a completely different task than~\cite{Jiang-AAAI-2015}, namely cross-domain object detection. To our knowledge, we are the first to study curriculum self-paced learning in the cross-domain object detection setting.

\noindent
{\bf Cross-Domain Object Detection.}
While domain adaptation has been extensively studied to address cross-domain classification, cross-domain object detection is a more challenging and less studied task, perhaps because it requires localizing each object in an image, in addition to identifying the corresponding object categories. \cite{Inoue-CVPR-2018} tackled the cross-domain weakly-supervised object detection task using a two-step progressive domain adaptation technique to fine-tune the detector trained on a source domain, while others~\citep{Chen-CVPR-2018,Khodabandeh-ICCV-2019,Raj-BMVC-2015,Saito-CVPR-2019,Shan-NC-2019,Zheng-CVPR-2020,Zhu-CVPR-2019} studied unsupervised cross-domain object detection methods. Although we include all these methods in our experiments, we consider fair only the comparison with the latter methods, as our approach falls in the same category of unsupervised methods. \cite{Raj-BMVC-2015} used subspace alignment, a domain adaptation method consisting of learning a mapping from the source distribution to the target one. \cite{Chen-CVPR-2018} argued that the gap between domains can be found both at the image level (illumination and style) and at the instance level (object size and overall appearance). Thus, they provided separate components to treat each case on top of a Faster Region-based Convolutional Neural Network (R-CNN)~\citep{Ren-NIPS-2015} detector. These components use a domain classifier and adversarial training to learn domain-invariant features. Different from~\cite{Raj-BMVC-2015} and~\cite{Chen-CVPR-2018}, we propose a curriculum self-paced learning approach to adapt the detector to the target domain.

\cite{Zhu-CVPR-2019} proposed a framework that focuses on aligning the local regions containing objects of interest. It consists of a region mining component, which finds relevant patches, and a region-level alignment component, which uses adversarial learning to align the image patches reconstructed from the features of the selected regions. \cite{Khodabandeh-ICCV-2019} proposed a robust framework which takes into consideration the generated labels of the target domain to retrain the detector on both domains. The robustness is defined against mistakes in both object classification and localization. Thus, during retraining, the model can change labels and detection boxes, refining the noisy labels on the target domain. To improve the detections even further, the authors used a supplementary classification module that provides information about the target domain. Unlike \cite{Zhu-CVPR-2019} and \cite{Khodabandeh-ICCV-2019}, we propose a more simple and effective framework that gradually learns from noisy pseudo-labels, using an easy-to-hard approach.

\cite{Saito-CVPR-2019} introduced an object detection framework that performs both strong local alignment and weak global alignment. Strong local alignment is obtained using a fully convolutional network with one-dimensional kernels as a local domain classifier trained to focus on local features. For the weak global feature alignment, the authors trained a domain classifier to ignore easy-to-classify examples while focusing on the more difficult ones, with respect to the domain classification. The reason behind this approach is that easy-to-classify target examples are far from the source in the feature space, while the harder examples are closer to the source. Different from~\cite{Saito-CVPR-2019}, we do not use a domain classifier to determine which samples are easy and which are difficult. Instead, we estimate the difficulty at the image level by computing the number of detected objects divided by their average bounding box area. This gives us a measure of difficulty from a different perspective, that of the object detector (not the one of the domain classifier). In our case, the object detector has higher accuracy for the easy image samples versus the difficult image samples.

\cite{Shan-NC-2019} proposed a multi-module framework consisting of a pixel-level domain adaptation module based on Cycle-GAN and a feature-level domain adaptation module based on Faster R-CNN. The pixel-level alignment is achieved by using a traditional generator-discriminator approach, with a loss function to ensure cycle consistency. In comparison, our method is a simple and straightforward combination of modules, adversarial domain adaptation and curriculum self-paced learning, stacked on top of a traditional Faster R-CNN baseline. We use Cycle-GAN to transfer from the source training set to the target set, thus generating additional training (labeled) information with similar style to the target domain. We then extract pseudo-labels from an already more trustworthy detector, and fine-tune it through curriculum self-paced learning. Our novel idea is that fine-tuning can be done in a meaningful, not random, order, which is defined by our measure of image difficulty.

\cite{Zheng-CVPR-2020} introduced a Graph-induced Prototype Alignment method to perform domain alignment for each object category through class prototypes. They designed the approach specifically for a two-stage detector, Faster R-CNN, performing the alignment in two stages. In the first stage, a relation graph is constructed to aggregate features at the instance level by considering both the location and the size of the object proposals. In the second stage, the information contained in various instances is gathered into class-level prototypes, thus enabling category-level domain alignment. Unlike~\cite{Zheng-CVPR-2020}, our method is not tightly coupled with the Faster R-CNN detector, being a generic domain adaption technique applicable to any object detector. We confirm this statement by showing that our domain adaptation method is also useful for a RetinaNet \citep{Lin-ICCV-2017} detector.
Moreover, our approach only consists in a training algorithm, not adding any additional branches to the object detector, as~\cite{Zheng-CVPR-2020}. Hence, different from~\cite{Zheng-CVPR-2020}, our method does not bring any computational overhead during inference.

\section{Method}
\label{sec_method}

\begin{figure*}[!t]
\begin{center}
\includegraphics[width=0.8\linewidth]{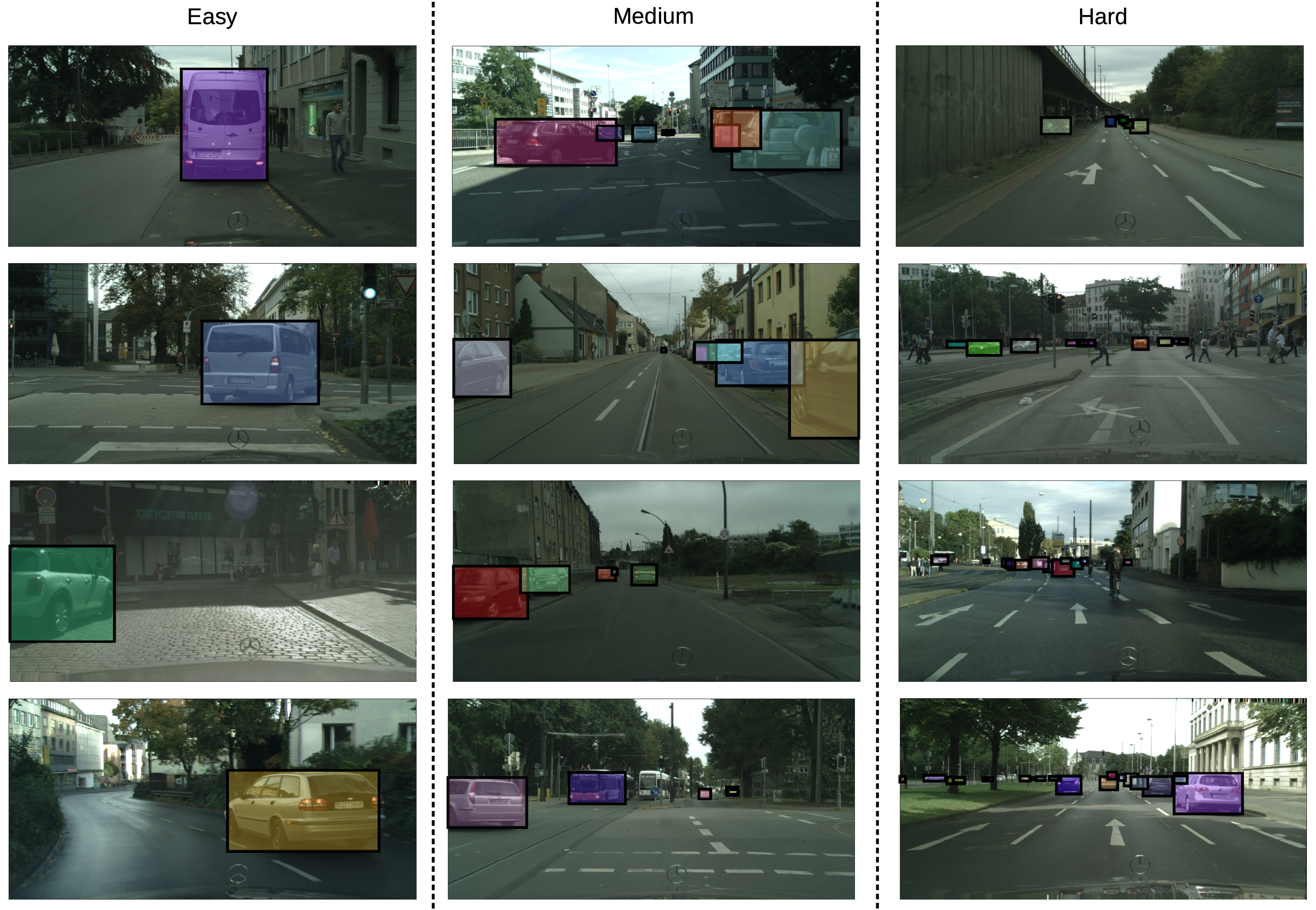}
\end{center}
\vspace*{-0.3cm}
\caption{Examples divided into easy, medium and hard batches according to the proposed difficulty metric. Easy images contain few cars that are closer to the camera, while hard images contain many cars that are father away from the camera. Best viewed in color.}
\label{fig_examples}
\end{figure*}

Domain adaptation is a fervent topic, many papers on object detection already taking advantage of adaptation methods to align models trained across domains~\citep{Chen-CVPR-2018,Khodabandeh-ICCV-2019,Saito-CVPR-2019,Shan-NC-2019,Zheng-CVPR-2020,Zhu-CVPR-2019}. The same consideration has been granted to self-supervised learning techniques, in which exploiting reliable pseudo-labels to improve classification has been already investigated~\citep{Doersch-ICCV-2015,Misra-ECCV-2016,Wei-CVPR-2018}. In this work, however, we aim at evaluating a model which incorporates a domain adaptation method based on style transfer using GAN-like preprocessing in conjunction with a self-paced learning method based on difficulty-wise curriculum learning provided by the difficulty metric proposed in Section \ref{sec:components}.

The general principle of the easy-to-hard training strategies, known as \emph{curriculum learning}~\citep{Bengio-ICML-2009}, stems from the fact that human beings learn better when they receive easy examples first, with gradually more complex concepts being introduced later. \cite{Bengio-ICML-2009} proved the effectiveness of this learning strategy for neural networks as well. Inspired by~\cite{Bengio-ICML-2009}, we propose a novel method to apply curriculum learning on object detection, replacing the random sampling during self-paced training. 
At this point, an important question arises: ``How do we define the difficulty of detecting objects in an image?'' Few different solutions have been proposed to address this problem~\citep{Ionescu-CVPR-2016,Soviany-SYNASC-2018,Wang-ICPR-2018,Zhang-IJCV-2019}. Different from these works, we propose a difficulty metric that can easily be computed based on the output bounding boxes provided by our object detector, eliminating the need for additional model components, e.g.~building an image difficulty regressor~\citep{Ionescu-CVPR-2016}. We provide a comparison of various difficulty metrics in Section~\ref{sec_experiments}, showing that our metric produces superior results.

We show that fine-tuning the model with pseudo-labels increases performance. Nonetheless, in our experiments, we found that it is more impactful when pseudo-labels are used on top of a ``warmed-up'' model, because, in this way, our confidence in the generated labels will be higher. Our aim is building a simple and efficient method which can be used together with almost any other domain adaptation strategy, so we did not alter the architecture of the standard Faster R-CNN~\citep{Ren-NIPS-2015} detector, nor used any other complex adaptation strategy. In order to reduce pseudo-labeling noise (increasing the performance by as much as possible) before applying fine-tuning on real target images, we translate the source images to the target domain using Cycle-GAN, then train on the resulting images together with the original source training set, thus warming-up the model before curriculum self-paced learning.

In the rest of this section, we briefly present the components employed in our framework and detail our algorithm.

\subsection{Components}
\label{sec:components}

\noindent
{\bf Object Detector}. As most related articles, we apply our domain adaptation framework on top of a Faster R-CNN object detector~\citep{Ren-NIPS-2015}, thus facilitating a fair comparison with state-of-the-art methods. Faster R-CNN is one of the state-of-the-art region-based deep detection models. It is a two-stage object detector which improves Fast R-CNN~\citep{Girshick-ICCV-2015} by introducing a Region Proposal Network. In order to select the right regions of interest, it uses a fully convolutional network that can predict object bounds at every location. The selected regions are then provided as an input to the Fast R-CNN model, which gives the final detection results. Our approach is not tied to Faster R-CNN, in theory, being applicable to any object detector. To demonstrate this property, we show that our domain adaptation approach can successfully adapt a RetinaNet~\citep{Lin-ICCV-2017} object detector to the target domain.

\noindent
{\bf Cycle-GAN.} Cycle-GAN~\citep{Zhu-ICCV-2017} is a generative model  performing image translation between two domains without requiring paired images for training. It learns the relevant features and the translation mapping by using cycle consistency, constraining the model so that translating from one domain to another and back again must reach the starting point. In a preliminary evaluation, we observed that Cycle-GAN produces better translations compared with a standard GAN, indicating that cycle consistency is indeed useful.

\noindent
{\bf{Difficulty Metric.}} \cite{Ionescu-CVPR-2016} observed that images containing many small objects are more difficult than images with few large objects. Thus, we could compute an image difficulty score as the number of detected objects divided by their average bounding box area. Given a set of $n$ bounding box detections $B = \{b_1, b_2, ..., b_n \}$ in an image $I$, where a detection $b_i$ is composed of a tuple $(x_i, y_i, w_i, h_i)$ representing the coordinates of the top left corner $(x_i,y_i)$, the width and the height of the bounding box, we define our difficulty scoring function $S$ as follows:
\begin{equation}\label{eq_difficulty}
S(I,B) = \frac{n}{\frac{1}{n}\sum_{i=1}^n w_i \cdot h_i}=\frac{n^2}{\sum_{i=1}^n w_i \cdot h_i}.
\end{equation}

This method is effective in our case, because it computes difficulty as a function of the detected instances. To avoid the degenerate case of an image without detections, we assign a very high score to it. In other words, we consider images without any detected objects as \textit{hard}.
More general difficulty measures, such as the one proposed by~\cite{Ionescu-CVPR-2016}, take into consideration the entire image, which includes the background and unlabeled objects belonging to classes not included in the evaluation, e.g. trees or buildings. Our metric is more focused and does not require any additional neural models.

We should notice that our difficulty metric is automatically computed via the object detector. Hence, the mistakes of the object detector can also propagate to our difficulty metric. Although we acknowledge this problem regarding the proposed metric, we observe that, in most cases, the difficulty scores are relevant (see Figure~\ref{fig_examples}). Additionally, we show $(i)$ the effectiveness of the proposed metric in the context of curriculum self-paced learning (see Table~\ref{tab_results}) and $(ii)$ the correlation of the induced easy, medium and difficult batches with respect to the performance level of the object detector (see Table~\ref{tab_difficulty_results}).

\subsection{Algorithm}

\begin{algorithm}[!t]
\small{
\caption{Our cross-domain object detection algorithm\label{alg_cross_obj_det}}

\SetKwInput{KwData}{Input}
\KwData{}

\ShowLn$X_s$ -- the source data set of samples;

\ShowLn$Y_s$ -- the ground-truth labels for source data $X_s$;


$X_t$ -- the target training set of unlabeled samples;

$X_t^{(test)}$ -- the target test set, where $X_t \cap X_t^{(test)} = \emptyset$;



$S$ -- a difficulty scoring function, e.g.~Equation~\eqref{eq_difficulty};

$k$ -- the number of batches to split by difficulty;

\BlankLine
\SetKwInput{KwData}{Notations}
\KwData{}

$D$ -- an object detector, e.g.~Faster R-CNN;

$T$ -- an image translation model, e.g.~Cycle-GAN;

$\tilde{X}_s$ -- the generated images with target domain style;

$\tilde{Y}_t$ -- the pseudo-labels for target data $X_t$;

\BlankLine
\SetKwInput{KwData}{Computation}
\KwData{}

$T \leftarrow train(T / X_s, X_t)$\;

$\tilde{X}_s \leftarrow T(X_s)$\;


$D \leftarrow train(D / (X_s,Y_s) \cup (\tilde{X_s}, Y_s))$\;


\For{$i \leftarrow 1,k$}
{
	$\tilde{Y}_t \leftarrow D(X_t)$\;

    $X_t^{(1),...,(k)}, \tilde{Y}_t^{(1),...,(k)} \leftarrow split(X_t, \tilde{Y}_t, k, S)$\;
	$D \leftarrow train(D/ \bigcup_{j=1}^i (X_t^{(j)}, \tilde{Y}_t^{(j)})$\;
}

$\mathcal{B} \leftarrow D(X_t^{(test)})$\;

\BlankLine
\SetKwInput{KwData}{Output}
\KwData{}

$\mathcal{B}$ -- the set of predicted bounding boxes.
}
\end{algorithm}

We next explain our algorithm, as illustrated in Figure~\ref{fig_pipeline} and formally presented in Algorithm~\ref{alg_cross_obj_det}. Our algorithm is divided into two phases: the first one for the warm-up of the detector $D$ (steps 11-13) and the second for the self-paced refinement (steps 14-17).

In our warm-up phase, we randomly sample a subset from both source and target sets, training a Cycle-GAN (step 11) to generate a set of samples $\tilde{X}_s$ (step 12) with an appearance similar to the target, but with the labels $Y_s$ inherited from the source. We keep the labels $Y_s$ for the generated images $\tilde{X}_s$ based on the assumption that Cycle-GAN transfers the style (object appearance, background), e.g.~from video game frames to real photos, while preserving locations and classes of objects. Using the samples generated by Cycle-GAN, we perform a traditional supervised training on the source domain using an out-of-the-box Faster R-CNN model (step 13). By mixing samples from the source domain with samples generated by Cycle-GAN, we produce a model that favors the alignment between the two domains, helping the self-paced learning on the unlabeled pristine target data set $X_t$.

The second phase is an iterative process described in steps 14 to 17 in Algorithm~\ref{alg_cross_obj_det}. At each iteration $i$, we first apply the current object detector $D$ on the target samples $X_t$ to produce the pseudo-labels $\tilde{Y}_t$ (step 15). The target samples are then ranked according to the proposed difficulty metric and divided into $k$ batches difficulty-wise (step 16, inside the $split$ function), i.e.~according to Equation \eqref{eq_difficulty}. Finally, the first $i$ batches, starting from those ranked as easier, are used for training the object detector $D$ (step 17). The curriculum self-paced learning process is repeated until eventually the whole target set has been included in the training process. It is important to note that only the high confidence detections have been taken into consideration, performing a threshold-based selection.



The intuition behind the usage of this curriculum fine-tuning approach over the standard random one relies on the simple fact that pseudo-labels for easier samples are typically more accurate. By using less difficult samples first, we can reduce the domain gap without learning too many wrongly detected objects. In this way, most pseudo-labels, even those of the harder samples, will be trustworthy, leading to higher performance after the final retraining step.

\section{Experiments}
\label{sec_experiments}

\subsection{Data Sets}

Following the methodology of previous studies~\citep{Chen-CVPR-2018,Khodabandeh-ICCV-2019}, we apply our method on two street scenes data set pairs, Sim10k$\rightarrow$Cityscapes and KITTI$\rightarrow$Cityscapes, considering only their common class, i.e.~\emph{car}. 
Sim10k~\citep{Johnson-Roberson-ICRA-2017} is a computer-generated data set of 10,000 images with traffic scenes, which we use as the source for our simulated-to-real domain transfer. KITTI~\citep{Geiger-IJRR-2013} is another driving data set consisting of 7,481 real training images that we use as source in our experiments which involve adaptation between two real data sets. Cityscapes~\citep{Cordts-CVPR-2016} contains 2,945 training images and 500 validation images of urban scenes. In our experiments, we use the training set (without ground-truth labels) for self-paced learning and the validation set for testing and evaluation.

In addition, we evaluate our method on two other pairs of data sets, PASCAL VOC 2007$\rightarrow$Clipart1k and PASCAL VOC 2007+2012$\rightarrow$Clipart1k, both having 20 object categories in common. PASCAL VOC 2007~\citep{Pascal-VOC-2007} and PASCAL VOC 2012~\citep{Pascal-VOC-2012} are well-known data sets, the former consisting of 9,963 images and the latter consisting of 11,540 images. Clipart1k~\citep{Inoue-CVPR-2018} contains 1,000 images that contain the same 20 categories as PASCAL VOC 2007 and 2012. We used 500 images (without ground-truth labels) for training and the other 500 images for testing.

\subsection{Experimental Setup}

\noindent
{\bf Evaluation Measures.} The performance of object detectors on a class of objects is typically evaluated using the Average Precision (AP), which is based on the ranking of detection scores~\citep{Pascal-VOC-2010}. We thus report the AP on each class and the  mean Average Precision (mAP) over all classes. The AP score is given by the area under the precision-recall (PR) curve for the detected objects. The PR curve is constructed by first mapping each detected bounding box to the most-overlapping ground-truth bounding box, according to the Intersection over Union (IoU) measure, but only if the IoU is higher than $0.5$~\citep{Everingham-IJCV-2015}. Then, the detections are sorted in decreasing order of their scores. Precision and recall values are computed each time a new positive sample is recalled. The PR curve is given by plotting the precision and recall pairs as lower-scored detections are progressively included.

To automatically quantify the quality and realism of artificially generated images (by Cycle-GAN), we employ the Inception Score~\citep{Salimans-NIPS-2016}, which is typically computed on five runs, with 10,000 generated images in each run. We note that a higher Inception Score (IS) indicates better performance.

\noindent
{\bf Baselines.}
In order to show the relevance of our approach, we compare our results with several state-of-the-art methods~\citep{Chen-CVPR-2018,Inoue-CVPR-2018,Khodabandeh-ICCV-2019,Saito-CVPR-2019,Shan-NC-2019,Zheng-CVPR-2020,Zhu-CVPR-2019}. Further results of previous studies can be consulted in these very recent works~\citep{Chen-CVPR-2018,Inoue-CVPR-2018,Khodabandeh-ICCV-2019,Saito-CVPR-2019,Shan-NC-2019,Zheng-CVPR-2020,Zhu-CVPR-2019}. 
We also include a domain adaptation model based solely on Cycle-GAN in our comparative experiments. Along with the best scores reported by each of these state-of-the-art methods, we also include the baseline detection models (without adaptation), observing the absolute gain in performance provided by each domain adaptation method with respect to the corresponding baseline.

\noindent
{\bf Implementation Details.}
We employ Faster R-CNN~\citep{Ren-NIPS-2015} based on the ResNet-50 \citep{He-CVPR-2016} backbone as our first object detector, just as~\citep{Chen-CVPR-2018,Khodabandeh-ICCV-2019,Saito-CVPR-2019,Shan-NC-2019,Zhu-CVPR-2019}. We use the same backbone for our second detector, RetinaNet~\citep{Lin-ICCV-2017}.
We use the PyTorch~\citep{Paszke-NIPSAutodiff-2017} implementations of Faster R-CNN and RetinaNet from~\citep{Massa-Web-2018} with weights pre-trained on ImageNet~\citep{Russakovsky-IJCV-2015}. We train each object detector for a number of 50,000 iterations, using adaptive learning rates. At the end of the training, we generate the pseudo-labels and apply self-paced learning for 500 iterations, with new training labels being generated at every 100 iterations. In the curriculum self-paced learning setup, we use easy images for the first 50 iterations, easy and medium images for the next 50 iterations, then the whole data set (including easy, medium and hard images) for the remaining iterations. The number of batches used in Algorithm~\ref{alg_cross_obj_det} is $k=3$. We perform image translation using Cycle-GAN~\citep{Zhu-ICCV-2017}\footnote{https://github.com/arnab39/cycleGAN-PyTorch}. We train a Cycle-GAN for 200 epochs on each data set pair. We use the same parameters in all our experiments, avoiding overfitting in hyperparameter space on individual data sets. All results are averaged over three runs.

\subsection{Unpaired Image Translation Results}

\begin{table}[!t]
\caption{Inception Scores (IS) of images generated by Cycle-GAN for Sim10k$\rightarrow$Cityscapes and KITTI$\rightarrow$Cityscapes in comparison with the Inception Scores of real images from Sim10k, KITTI and Cityscapes. Higher IS values represent better quality images.\label{tab_cycle_gan}}
\setlength\tabcolsep{3.8pt}
\begin{center}
\begin{tabular}{|l|c|c|}
\hline
Data Set                        & Images Type      & IS\\
\hline
\hline
Sim10k                          & real      & $4.68$\\
KITTI                           & real      & $3.77$\\
Cityscapes                      & real      & $3.70$\\
Sim10k$\rightarrow$Cityscapes   & translated by Cycle-GAN   & $4.32$\\
KITTI$\rightarrow$Cityscapes    & translated by Cycle-GAN   & $3.36$ \\
\hline
\end{tabular}
\end{center}
\vspace{-0.2cm}
\end{table}

\begin{table}[!t]
\caption{AP and mAP scores for easy, medium and hard image batches, provided by Faster R-CNN trained on original source images and on images translated by Cycle-GAN. Results are reported for Sim10k$\rightarrow$Cityscapes, KITTI$\rightarrow$Cityscapes and PASCAL VOC 2007$\rightarrow$Clipart1k benchmarks.\label{tab_difficulty_results}}
\setlength\tabcolsep{3.8pt}
\begin{center}
\begin{tabular}{|l|c|c|c|}
\hline
Data Set                        & Easy      & Medium    & Hard \\
\hline
\hline
Sim10k$\rightarrow$Cityscapes   & $44.43$   & $43.51$   & $36.90$ \\
KITTI$\rightarrow$Cityscapes    & $40.70$   & $40.05$   & $38.08$ \\
PASCAL VOC 2007$\rightarrow$Clipart1k    & $29.21$   & $18.87$   & $13.00$ \\
\hline
\end{tabular}
\end{center}
\vspace{-0.2cm}
\end{table}

\begin{table*}[!t]
\caption{AP and mAP scores (in $\%$) of several Faster R-CNN models trained using different state-of-the-art domain adaptation methods~\citep{Chen-CVPR-2018,Inoue-CVPR-2018,Khodabandeh-ICCV-2019,Saito-CVPR-2019,Shan-NC-2019,Zheng-CVPR-2020,Zhu-CVPR-2019} versus a Faster R-CNN model trained using our domain adaptation approach based on curriculum self-paced learning. All domain-adapted (DA) methods include images without ground-truth labels from the target domain. Faster R-CNN baselines (B) without adaptation, i.e. trained only on source, are also included to point out the absolute gain of each domain adaptation technique, with respect to the corresponding baseline. Faster R-CNN models trained on target domain (TD) images with ground-truth label are included as indicators of possible upper bounds of the AP scores. Results are reported for Sim10k$\rightarrow$Cityscapes, KITTI$\rightarrow$Cityscapes, PASCAL VOC 2007$\rightarrow$Clipart1k and PASCAL VOC 2007+2012$\rightarrow$Clipart1k benchmarks. The best scores and the highest absolute gains are highlighted in bold.\label{tab_results}}
\setlength\tabcolsep{1.4pt}
\begin{center}
\begin{tabular}{|l|l|c|c|c|c|}
\hline
Faster R-CNN            & Train Data        & Sim10k$\rightarrow$City       & KITTI$\rightarrow$City & VOC07$\rightarrow$Clipart  & VOC07+12$\rightarrow$Clipart\\
\hline
\hline
B~\citep{Chen-CVPR-2018}          & Source                      & $30.12$           & $30.20$       & -         & - \\
B~\citep{Inoue-CVPR-2018}         & Source                        & -                 & -             & -         & $26.20$ \\
B~\citep{Khodabandeh-ICCV-2019}   & Source                        & $31.08$           & $31.10$       & -         & - \\
B~\citep{Saito-CVPR-2019}         & Source                        & $34.60$           & -             & -         & - \\
B~\citep{Shan-NC-2019}            & Source                        & $30.10$           & $30.20$       & -         & - \\
B~\citep{Zheng-CVPR-2020}        & Source                        & $34.60$           & $37.60$       & -         & - \\
B~\citep{Zhu-CVPR-2019}           & Source                        & $33.96$           & $37.40$       & -         & - \\
B (ours)                         & Source                        & $30.67$           & $31.52$       & $18.73$   & $26.14$\\
\hline
DA~\citep{Chen-CVPR-2018}        & Source+Target (no labels)   & $38.97~(+8.85)$   & $38.50~(+8.30)$       & -         & - \\
DA~\citep{Inoue-CVPR-2018}       & Source+Target            & -                 & -                     & -         & $34.90~(8.70)$\\
DA~\citep{Khodabandeh-ICCV-2019} & Source+Target (no labels)   & $42.56~(+11.48)$  & $42.98~(+11.88)$      & -         & - \\
DA~\citep{Saito-CVPR-2019}       & Source+Target (no labels)   & $40.70~(+5.80)$   & -                     & -         & - \\
DA~\citep{Shan-NC-2019}          & Source+Target (no labels)   & $39.60~(+9.50)$   & $41.80~(+11.60)$      & -         & - \\
DA~\citep{Zheng-CVPR-2020}       & Source+Target (no labels)   & $47.60~(+13.00)$   & $\mathbf{47.90}~(+10.30)$       & -         & - \\
DA~\citep{Zhu-CVPR-2019}         & Source+Target (no labels)   & $43.02~(+9.06)$   & $42.50~(+5.10)$       & -         & - \\
DA (Cycle-GAN)                  & Source+Target (no labels)   & $41.53~(+10.86)$  & $38.72~(+7.20)$       & $21.31~(+2.58)$       & $35.60~(+9.46)$ \\
\hline
DA (ours)                         & Source+Target (no labels)   & $\mathbf{47.68}~(+\mathbf{17.01})$  & $43.86~(+\mathbf{12.34})$ & $\mathbf{27.64}~(+\mathbf{8.91})$ & $\mathbf{37.83}~(+\mathbf{11.69})$\\
\hline
TD~\citep{Inoue-CVPR-2018}         & Target                       & -                 & -             & -         & $50.00$\\
TD~\citep{Khodabandeh-ICCV-2019}   & Target                        & $68.10$           & $68.10$      & -         & -\\
TD~\citep{Saito-CVPR-2019}         & Target                        & $53.10$           & $53.10$      & -         & -\\
TD (ours)                         & Target                        & $62.73$           & $62.73$      & $33.89$   & $33.89$\\
\hline
\end{tabular}%
\end{center}
\end{table*}

It is important to note that the quality of the images translated by Cycle-GAN can indirectly influence the effectiveness of our domain adaptation approach. Therefore, we analyze the quality of the generated samples for two data set pairs, Sim10k$\rightarrow$Cityscapes and KITTI$\rightarrow$Cityscapes. In Table~\ref{tab_cycle_gan}, we compare the Inception Scores of the generated samples with the Inception Scores of the real data samples in Sim10k, KITTI and Cityscapes. First, we observe that the Inception Scores of the generated images are in the same range as the Inception Scores of the real images. Nonetheless, we should emphasize that the Inception Scores of the translated images cannot be higher than the Inception Scores of the real images from the source data set. As Sim10k exhibits a higher IS than KITTI, it seems reasonable to have a higher IS for the images generated from Sim10k as source than for the images generated from KITTI. Perhaps, this phenomenon could also explain why the Sim10k$\rightarrow$Cityscapes domain adaption gives a higher absolute performance gain than the KITTI$\rightarrow$Cityscapes adaptation (see Table~\ref{tab_results}). In summary, the results reported in Table~\ref{tab_cycle_gan} indicate that the quality and realism of the samples generated by Cycle-GAN are adequate.

\subsection{Preliminary Results}

We conduct a preliminary set of experiments to validate our hypothesis stating that the number of objects divided by their average bounding box area is a good measure of image difficulty in the context of object detection. We first train the Faster R-CNN on original source images and on images translated by Cycle-GAN. The model is thus already adapted to the target domain and should provide more reliable labels on real target images. We next apply the model on target domain images and we divide the images into $k=3$ batches, in increasing order of difficulty. We provide the corresponding results on all benchmarks in Table~\ref{tab_difficulty_results}. We note that the AP scores on the easy batch of images are higher than the AP scores on the medium batch. We observe the same behavior on the medium batch with respect to the hard batch. In conclusion, the empirical results presented in Table~\ref{tab_difficulty_results} confirm our hypothesis. We can thus apply the proposed difficulty measure in our curriculum self-paced learning approach.

\begin{table*}[!t]
\caption{AP and mAP scores (in $\%$) of various ablated versions of our framework versus our full framework based on Faster R-CNN. Results for the baseline Faster R-CNN, trained only on source (S) data, and the in-domain Faster R-CNN, trained on target (T) data with ground-truth labels, are also included for comparison.\label{tab_ablation}}
\setlength\tabcolsep{1.8pt}
\begin{center}
\begin{tabular}{|l|c|c|c|c|}
\hline
Train Data          & Sim10k$\rightarrow$City   & KITTI$\rightarrow$City & VOC07$\rightarrow$Clipart & VOC07+12$\rightarrow$Clipart\\
\hline
\hline
Source                                               & $30.67$           & $31.52$       & $18.73$       & $26.14$\\
\hline
Source+Target (self-paced)                                & $34.39$           & $35.64$       & $20.33$       & $30.76$\\
Source+Target (curriculum)                                & $35.80$           & $37.02$       & $21.11$       & $31.69$\\
Source+Source$\rightarrow$Target (Cycle-GAN)                  & $41.53$           & $38.72$       & $21.31$       & $30.43$\\
Source+Source$\rightarrow$Target (Cycle-GAN)+Target (self-paced) & $46.84$           & $41.52$       & $25.53$       & $35.60$\\
Source+Source$\rightarrow$Target (Cycle-GAN)+Target (curriculum) & $47.68$           & $43.86$       & $27.64$       & $37.83$\\
\hline
Target                                               & $62.73$           & $62.73$       & $33.89$       & $33.89$\\
\hline
\end{tabular}
\end{center}
\end{table*}

\subsection{Cross-Domain Detection Results}

We compare our domain adaptation method with several state-of-the-art approaches~\citep{Chen-CVPR-2018,Inoue-CVPR-2018,Khodabandeh-ICCV-2019,Saito-CVPR-2019,Shan-NC-2019,Zheng-CVPR-2020,Zhu-CVPR-2019} and a domain adaptation model based on Cycle-GAN. We provide the comparative object detection results on Sim10k$\rightarrow$Cityscapes, KITTI$\rightarrow$Cityscapes, PASCAL VOC 2007$\rightarrow$Clipart1k and PASCAL VOC 2007+2012$\rightarrow$Clipart1k benchmarks in Table~\ref{tab_results}. 

First, we note that each state-of-the-art method is applied on top of a slightly different Faster R-CNN baseline (trained on source only). While three methods~\citep{Saito-CVPR-2019,Zheng-CVPR-2020,Zhu-CVPR-2019} start from somewhat better Faster R-CNN versions, our Faster R-CNN baseline gives similar AP scores to the Faster R-CNN baselines used in~\citep{Chen-CVPR-2018,Khodabandeh-ICCV-2019,Shan-NC-2019}. Since the baselines are not equally good, we report the absolute gains with respect to the corresponding baseline along with the AP scores, for a more fair comparison between the domain adaptation methods. 

On Sim10k$\rightarrow$Cityscapes, we obtain the best AP score ($47.68\%$) among all methods, as well as the largest improvement over the corresponding baseline ($17.01\%$). In terms of AP, the second best result, reported by~\cite{Zheng-CVPR-2020}, is $0.08\%$ lower. The improvement of~\cite{Zheng-CVPR-2020} over their baseline is however much lower ($13.00\%$). Indeed, in terms of absolute gain over the corresponding baseline, our gain is $4.01\%$ higher than that of~\cite{Zheng-CVPR-2020}. We conclude that our domain-adaptation method attains significant improvements over the state-of-the-art methods on the Sim10k$\rightarrow$Cityscapes benchmark.

On KITTI$\rightarrow$Cityscapes, we obtain the second best AP score ($43.86\%$), which is $4.04\%$ lower than the top scoring method~\citep{Zheng-CVPR-2020} in literature. However, we should point out that~\cite{Zheng-CVPR-2020} started from a much better Faster R-CNN baseline. Hence, our absolute gain ($12.34\%$) is still higher than the absolute gain ($10.30\%$) of~\cite{Zheng-CVPR-2020}. In fact, two other methods \citep{Khodabandeh-ICCV-2019,Shan-NC-2019} from the recent literature also surpass~\cite{Zheng-CVPR-2020} in terms of absolute gain with respect to the corresponding baseline. The other methods from the recent literature~\citep{Chen-CVPR-2018,Saito-CVPR-2019,Zhu-CVPR-2019} attain lower AP scores as well as lower absolute gains. We conclude that our method is better than all other methods on KITTI$\rightarrow$Cityscapes, specifically in terms of the absolute gain with respect to the corresponding baseline.

To our knowledge, there are no previous results reported on PASCAL VOC 2007$\rightarrow$Clipart1k. However, we compare our approach with a strong domain adaptation model based on Cycle-GAN. Indeed, this model surpasses three state-of-the-art methods~\citep{Chen-CVPR-2018,Saito-CVPR-2019,Shan-NC-2019} on Sim10k$\rightarrow$Cityscapes and one state-of-the-art method~\citep{Chen-CVPR-2018} on KITTI$\rightarrow$Cityscapes. With an mAP of $27.64\%$ on PASCAL VOC 2007$\rightarrow$Clipart1k, our method surpasses the Faster R-CNN trained on source only by $8.91\%$ and the domain adaptation model based on Cycle-GAN by $6.33\%$. 

On PASCAL VOC 2007+2012$\rightarrow$Clipart1k, we compare our results with those reported by \cite{Inoue-CVPR-2018} and those of the domain adaptation model based on Cycle-GAN. Although our baseline Faster R-CNN and that of \cite{Inoue-CVPR-2018} obtain about the same results (just above $26\%$), there is a high difference between the mAP of $33.89\%$ attained by our Faster R-CNN trained directly on the target domain (with supervision) and the mAP of $50.00\%$ reported by \cite{Inoue-CVPR-2018} for their in-domain Faster R-CNN. Regarding the domain adapted models, we obtain a better mAP as well as a higher absolute gain than competing methods. In terms of AP, our improvement is of $2.93\%$. Interestingly, all the domain adapted models surpass our in-domain Faster R-CNN. This can be explained by the fact that the number of training images in Clipart1k is rather small (500) in comparison to the number of images in PASCAL VOC 2007+2012. We believe that the high number and the variety of samples in the source domain compensate for the fact that the samples come from a different domain.

\subsection{Ablation Study}

We conduct an ablation study to determine the benefits of each individual component in our framework. Table~\ref{tab_ablation} illustrates our results on all four cross-domain benchmarks, indicating the contribution of each component to the complete model in terms of AP and mAP scores, respectively.

\noindent
{\bf Source$\rightarrow$Target Translation with Cycle-GAN.} With respect to the Faster R-CNN baselines trained on source only, we gain $10.86\%$ on Sim10k$\rightarrow$Cityscapes, $7.22\%$ on KITTI$\rightarrow$Cityscapes, $2.58\%$ on PASCAL VOC 2007$\rightarrow$Clipart1k and $4.29\%$ on PASCAL VOC 2007+2012$\rightarrow$Clipart1k. We observe that training on images translated from KITTI to Cityscapes is less effective than from Sim10k to Cityscapes. We suspect that the gap between synthetic (Sim10k) and real data (Cityscapes) can be easier to bridge than the gap induced by different cameras and object sizes/view angles between data sets containing real scenes (KITTI and Cityscapes). As also mentioned earlier, another explanation for the less effective KITTI$\rightarrow$Cityscapes transfer is the lower Inception Score of the real KITTI images with respect to the real Sim10k images (see Table~\ref{tab_cycle_gan}).


\noindent
{\bf Self-Paced Learning.} We employ self-paced learning from pseudo-labels either on top of the model trained only on source data or on top of the model trained with additional data produced via Cycle-GAN translation. On all four data set pairs, we observe a consistent increase in performance of around $4\%$. Regarding the first three benchmarks, we observe that self-paced learning alone does not reach the performance gains of the Cycle-GAN adaptation approach, with a difference of $7.14\%$ on Sim10k$\rightarrow$Cityscapes, $3.1\%$ on KITTI$\rightarrow$Cityscapes and $0.98\%$ on PASCAL VOC 2007$\rightarrow$Clipart1k. Nonetheless, self-paced learning is superior to Cycle-GAN domain adaptation in the last benchmark. Furthermore, the accuracy improvements brought by self-paced learning are still visible on the models that are already trained using Cycle-GAN translation, supporting our decision to perform self-paced learning on top of Cycle-GAN adaptation.

\begin{figure*}[!t]
\begin{center}
\includegraphics[width=0.9\linewidth]{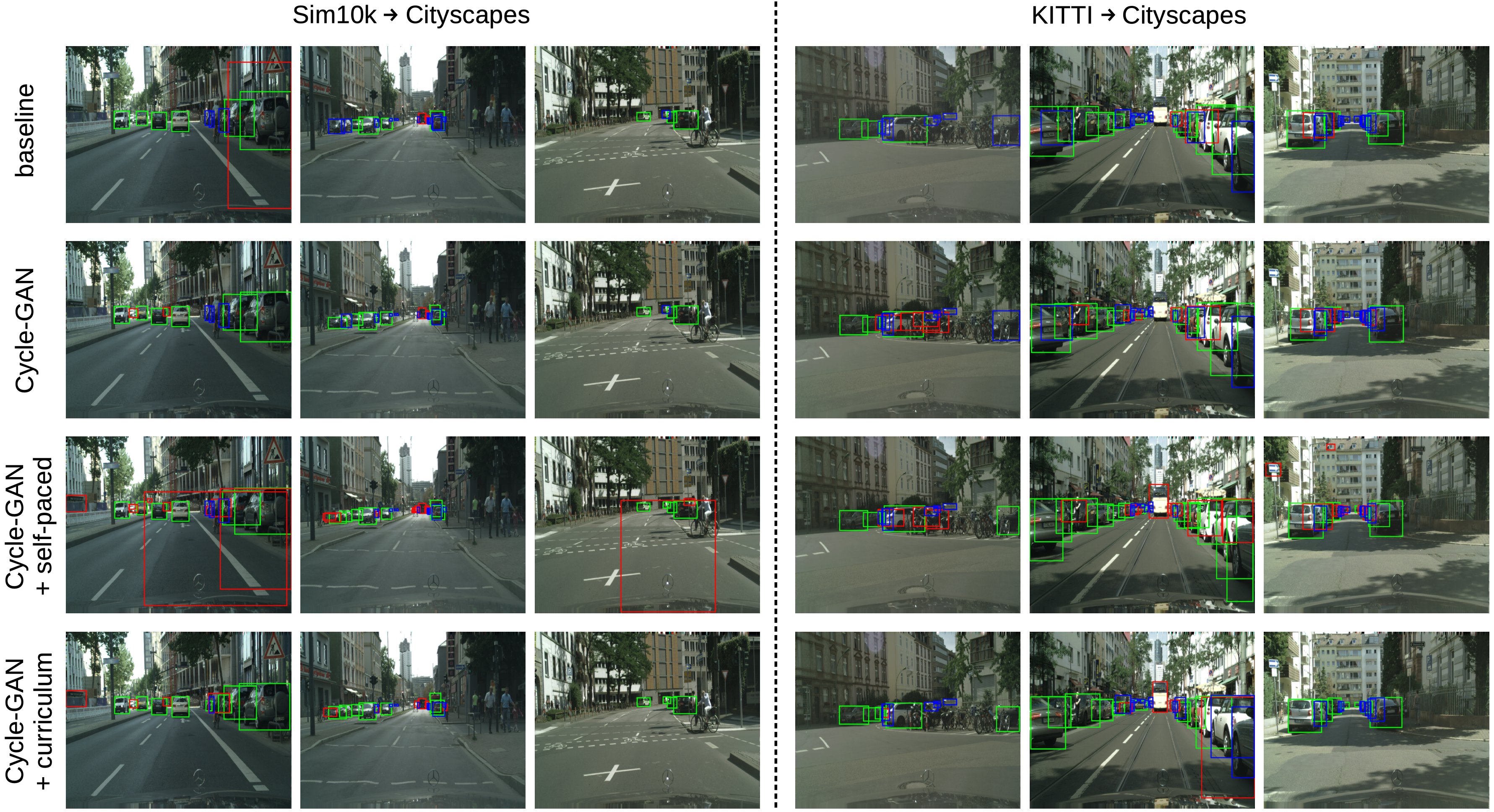}
\end{center}
\vspace*{-0.3cm}
\caption{Examples of detected cars provided by the baseline Faster R-CNN (first row) versus detections provided by two ablated versions of our framework (second and third rows) and our full domain adaptation framework based on Cycle-GAN and curriculum self-paced learning (fourth row). Samples are selected from Sim10k$\rightarrow$Cityscapes (first three columns) and KITTI$\rightarrow$Cityscapes (last three columns) experiments. Green bounding boxes represent correct detections; red bounding boxes represent false positives; blue bounding boxes represent false negatives. Best viewed in color.}
\label{fig_results}
\end{figure*}

\noindent
{\bf Curriculum Self-Paced Learning.} Our best results are not obtained, though, using basic self-paced learning, but using a curriculum learning approach in which the fine-tuning is conducted by gradually adding more difficult image batches. Our results show a typical gain of around $2\%$ over standard self-paced learning. Although curriculum self-paced learning alone surpasses the Cycle-GAN translation in only one cross-domain benchmark (PASCAL VOC 2007+2012$\rightarrow$Clipart1k) when applied on the baseline model (trained only on source), the complete framework, with all the components in place, provides state-of-the-art improvements in each and every case. Another benefit of our curriculum learning approach over the standard one is that it provides more stable results, and the results can be easily replicated under random initialization or different self-paced learning settings. 

\subsection{Qualitative Analysis}

Figure~\ref{fig_results} illustrates some typical detection results of the baseline Faster R-CNN versus our framework. Object detections provided by ablated versions of our framework are also included. In general, we observe that the domain-adapted models are able to detect more cars (depicted inside green bounding boxes in Figure~\ref{fig_results}), i.e. the number of false negatives (blue bounding boxes in Figure~\ref{fig_results}) is reduced. In the same time, the domain-adapted models give more false positives (red bounding boxes in Figure~\ref{fig_results}). It seems that the self-paced learning framework applied after Cycle-GAN adaptation (third row in Figure~\ref{fig_results}) has more false positives than the other domain adaptation methods (second and fourth rows in Figure~\ref{fig_results}).

\subsection{Varying the Number of Curriculum Batches}

\begin{figure}[!t]
\begin{center}
\includegraphics[width=1.0\linewidth]{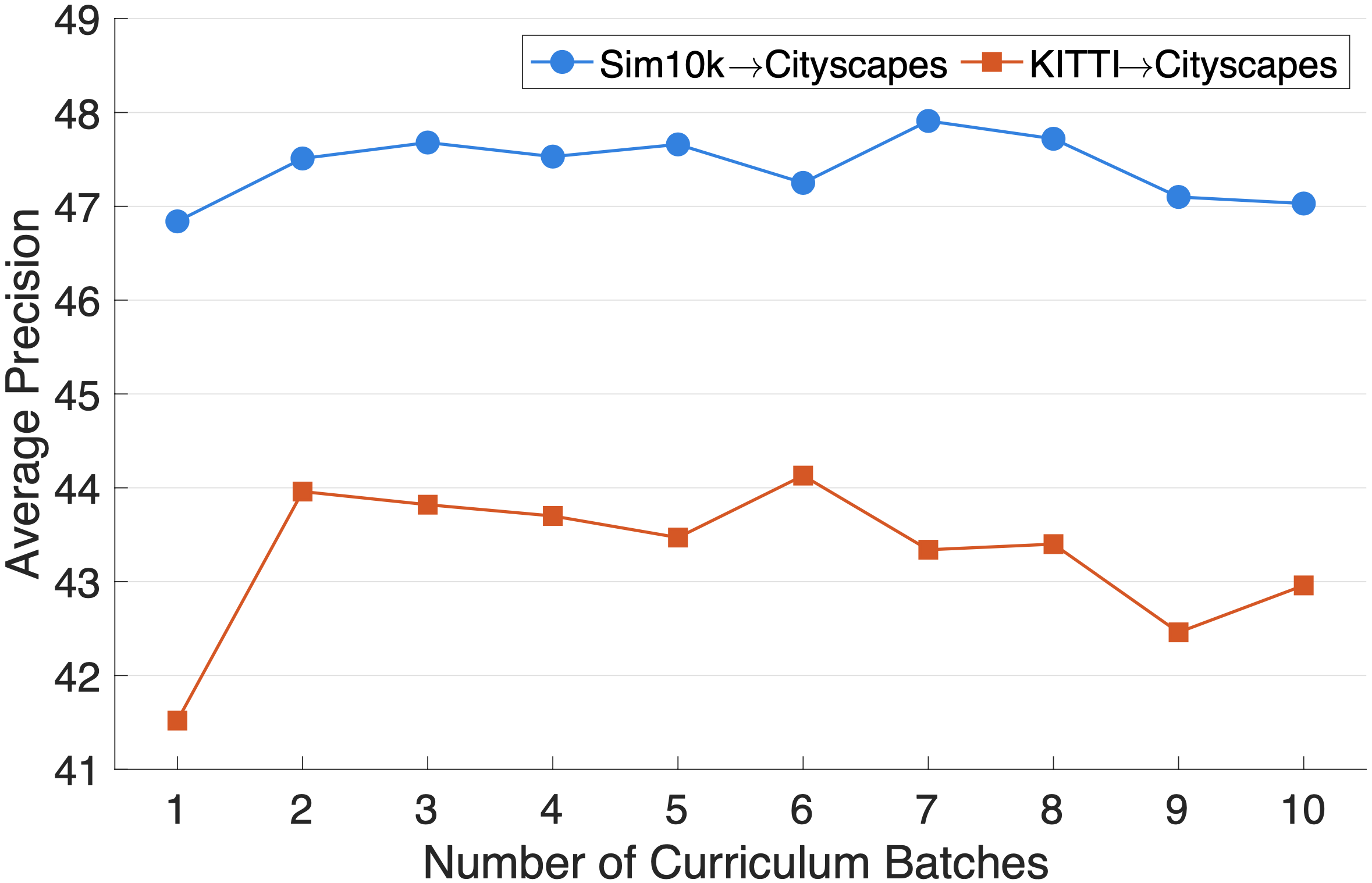}
\end{center}
\vspace*{-0.3cm}
\caption{AP scores (in $\%$) with different values for $\mathbf{k}$ (the number of batches) between $\mathbf{2}$ and $\mathbf{10}$, during curriculum self-paced learning on Sim10k$\rightarrow$Cityscapes (blue) and KITTI$\rightarrow$Cityscapes (red). Best viewed in color.}
\label{fig_num_batches}
\end{figure}

\begin{table*}[!t]
\caption{AP scores (in $\%$) of a state-of-the-art domain adaptation method~\citep{Chen-CVPR-2018} versus our domain adaptation approach based on curriculum self-paced learning, considering the exact same Faster R-CNN baseline (B). Both domain-adapted (DA) methods include images without ground-truth labels from the target domain. The Faster R-CNN baseline (B) without adaptation, i.e. trained only on source, is also included to point out the absolute gain of each domain adaptation technique. Results are reported for Sim10k$\rightarrow$Cityscapes and KITTI$\rightarrow$Cityscapes.\label{tab_results_rebase}}
\begin{center}
\begin{tabular}{|l|l|c|c|}
\hline
Faster R-CNN            & Train Data        & Sim10k$\rightarrow$City       & KITTI$\rightarrow$City\\
\hline
\hline
B~\citep{Chen-CVPR-2018}         & Source                      & $30.12$           & $30.20$ \\
\hline
DA~\citep{Chen-CVPR-2018}        & Source+Target (no labels)   & $38.97~(+8.85)$   & $38.50~(+8.30)$ \\
DA (ours) on top of B~\citep{Chen-CVPR-2018}    & Source+Target (no labels)   & ${45.98}~(+{15.31})$  & $42.89~(+{12.69})$ \\
\hline
\end{tabular}%
\end{center}
\vspace{-0.2cm}
\end{table*}

\begin{table*}[!t]
\caption{AP scores (in $\%$) of various ablated versions of our framework versus our full framework, considering two object detectors as baselines: RetinaNet and Faster R-CNN. Results are reported for Sim10k$\rightarrow$Cityscapes and KITTI$\rightarrow$Cityscapes.\label{tab_results_retina}}
\begin{center}
\begin{tabular}{|l|l|c|c|}
\hline
Detector        & Train Data                                                        & Sim10k$\rightarrow$City   & KITTI$\rightarrow$City \\
\hline
\hline
RetinaNet       & Source                                                            & $41.56$           & $39.35$ \\
\hline
RetinaNet       & Source + Target (self-paced)                                        & $47.37$           & $42.71$ \\
RetinaNet       & Source + Target (curriculum)                                        & $48.41$           & $43.87$ \\
RetinaNet       & Source + Source$\rightarrow$Target (Cycle-GAN)                     & $46.43$           & $42.00$ \\
RetinaNet       & Source + Source$\rightarrow$Target (Cycle-GAN) + Target (self-paced) & $49.81$           & $43.95$ \\
RetinaNet       & Source + Source$\rightarrow$Target (Cycle-GAN) + Target (curriculum) & $51.04$           & $46.68$ \\
\hline
RetinaNet       & Target                                                            & $66.03$           & $66.03$ \\
\hline
\hline
Faster R-CNN    & Source                                                            & $30.67$           & $31.52$ \\
\hline
Faster R-CNN    & Source + Target (self-paced)                                        & $34.39$           & $35.64$ \\
Faster R-CNN    & Source + Target (curriculum)                                        & $35.80$           & $37.02$ \\
Faster R-CNN    & Source + Source$\rightarrow$Target (Cycle-GAN)                     & $41.53$           & $38.72$ \\
Faster R-CNN    & Source + Source$\rightarrow$Target (Cycle-GAN) + Target (self-paced) & $46.84$           & $41.52$ \\
Faster R-CNN    & Source + Source$\rightarrow$Target (Cycle-GAN) + Target (curriculum) & $47.68$           & $43.86$ \\
\hline
Faster R-CNN    & Target                                                                & $62.73$           & $62.73$ \\
\hline
\end{tabular}
\end{center}
\vspace{-0.2cm}
\end{table*}

In the experiments presented so far, we fixed the number of curriculum batches to $k=3$ without particularly tuning this parameter. However, it is interesting to analyze the influence of $k$ on the performance level of the object detector. We thus present results on Sim10k$\rightarrow$Cityscapes and KITTI$\rightarrow$Cityscapes with various values for $k \in \{1,2,...,10 \}$ in Figure~\ref{fig_num_batches}. When $k=1$, curriculum self-paced learning becomes simple self-paced learning (without curriculum). The results reported in Figure~\ref{fig_num_batches} indicate that all AP scores for $k \in \{2,3,...,10\}$ are superior to the AP scores for $k=1$, indicating that curriculum is useful regardless of the number of batches. On Sim10k$\rightarrow$Cityscapes, the best AP score ($47.91\%$) is obtained for $k=7$, while on KITTI$\rightarrow$Cityscapes, the top AP score ($44.13\%$) is reported for $k=6$. Therefore, we conclude that our choice of fixing $k=3$ for all the other cross-domain object detection experiments is not optimal. Since the AP scores for $k \in \{2,3,...,8\}$ are fairly stable, setting $k=3$ does not seem such a bad choice in the end. On both data set pairs, we observe slight performance drops for $k \geq 9$, yet still higher than $k=1$.

\subsection{Changing the Baseline Faster R-CNN}

As it is hard to replicate the Faster R-CNN baseline results reported in competing works due to the unavailability of pre-trained models and the stochasticity of the optimization algorithm, the commonly agreed evaluation methodology in cross-domain object detection implies starting from a slightly different baseline. This statement is confirmed by works such as~\citep{Chen-CVPR-2018,Khodabandeh-ICCV-2019,Saito-CVPR-2019,Shan-NC-2019,Zheng-CVPR-2020,Zhu-CVPR-2019}. However, a completely fair evaluation in terms of AP can only be made by starting from the exact same baseline model. Since we were able to obtain the pre-trained Faster R-CNN baseline of~\cite{Chen-CVPR-2018}, we applied our cross-domain adaptation approach on top of their baseline, reporting the corresponding results in Table~\ref{tab_results_rebase}. We hereby note that our domain adaptation method outperforms the approach of~\cite{Chen-CVPR-2018} by a significant margin on both Sim10k$\rightarrow$Cityscapes and KITTI$\rightarrow$Cityscapes. Our absolute performance gains are within the same range as those reported in Table~\ref{tab_results} for our method (which was applied on top of our own Faster R-CNN baseline), confirming the significant improvements regardless of the Faster R-CNN baseline used as starting point.

\subsection{Applicability to Other Detectors (RetinaNet)}

In order to demonstrate that our cross-domain adaptation approach is not tied to the Faster R-CNN model, we perform domain adaption experiments considering a RetinaNet object detector as baseline. In Table~\ref{tab_results_retina}, we report the cross-domain results of RetinaNet, comparing the performance gains induced by each component in our framework on top of RetinaNet with the analogous gains on top of Faster R-CNN (the results for Faster R-CNN are copied from Table~\ref{tab_ablation} to improve readability). We observe that the RetinaNet baseline is significantly better than the Faster R-CNN baseline. Even so, our domain adaptation approach based on Cycle-GAN and curriculum self-paced learning brings significant improvements to the RetinaNet baseline, our absolute gains being $+9.48\%$ on Sim10k$\rightarrow$Cityscapes and $+7.33\%$ on KITTI$\rightarrow$Cityscapes. In summary, we conclude that our domain adaptation approach is applicable to at least two object detectors, namely Faster R-CNN and RetinaNet.

\begin{table*}[!t]
\caption{AP and mAP scores for easy, medium and hard image batches, provided by Faster R-CNN trained on original source images and on images translated by Cycle-GAN. Results are reported for Sim10k$\rightarrow$Cityscapes, KITTI$\rightarrow$Cityscapes and PASCAL VOC 2007$\rightarrow$Clipart1k benchmarks. The batches are obtained using three different difficulty estimation methods.\label{tab_suppl_difficulty_results}}
\begin{center}
\begin{tabular}{|l|c|c|c|c|}
\hline
Data Set                        & Difficulty Measure                    & Easy      & Medium    & Hard \\
\hline
\hline
Sim10k$\rightarrow$Cityscapes   & \#objects / average size (ours)               & $44.43$   & $43.51$   & $36.90$ \\
Sim10k$\rightarrow$Cityscapes   & image difficulty predictor~\citep{Ionescu-CVPR-2016}      & $42.01$   & $43.43$   & $43.36$ \\
Sim10k$\rightarrow$Cityscapes   & domain discriminator~\citep{Saito-CVPR-2019}      & $42.96$   & $40.76$   & $44.95$ \\
\hline
KITTI$\rightarrow$Cityscapes    & \#objects / average size (ours)              & $40.70$   & $40.05$   & $38.08$ \\
KITTI$\rightarrow$Cityscapes    & image difficulty predictor~\citep{Ionescu-CVPR-2016}      & $38.00$   & $39.88$   & $40.36$ \\
KITTI$\rightarrow$Cityscapes    & domain discriminator~\citep{Saito-CVPR-2019}      & $39.85$   & $37.81$   & $41.35$ \\
\hline
PASCAL VOC 2007$\rightarrow$Clipart1k    & \#objects / average size (ours)              & $29.21$   & $18.87$   & $13.00$ \\
PASCAL VOC 2007$\rightarrow$Clipart1k    & image difficulty predictor~\citep{Ionescu-CVPR-2016}      & $26.72$   & $21.02$   & $20.89$ \\
PASCAL VOC 2007$\rightarrow$Clipart1k    & domain discriminator~\citep{Saito-CVPR-2019}      & $23.84$   & $20.92$   & $25.43$ \\
\hline
\end{tabular}
\end{center}
\end{table*}

\begin{table*}[!t]
\caption{AP and mAP scores (in $\%$) of the self-paced learning method versus three curriculum self-paced learning methods. Each curriculum method is based on a different difficulty estimation approach. All self-paced learning methods are applied after domain adaptation with Cycle-GAN. Results are reported for Sim10k$\rightarrow$Cityscapes,  KITTI$\rightarrow$Cityscapes and PASCAL VOC 2007$\rightarrow$Clipart1k benchmarks. Best results are highlighted in bold.\label{tab_suppl_diff_ablation}}
\begin{center}
\begin{tabular}{|l|c|c|c|}
\hline
Sample selection for self-paced learning            & Sim10k$\rightarrow$City  & KITTI$\rightarrow$City & VOC$\rightarrow$Clipart\\
\hline
\hline

random                                                              & $46.84$    & $41.52$      & $25.53$ \\
image difficulty predictor~\citep{Ionescu-CVPR-2016}   & $47.02$    & $42.13$          & $26.61$ \\
domain discriminator~\citep{Saito-CVPR-2019}           & $45.76$    & $40.22$          & $25.28$ \\
\#objects / average size (ours)       & $\mathbf{47.68}$    & $\mathbf{43.86}$    & $\mathbf{27.64}$\\
\hline
\end{tabular}
\end{center}
\end{table*}

\subsection{Comparison with Other Difficulty Metrics}

In this work, we proposed to measure difficulty based on the number of detected objects divided by their average bounding box area, as defined in Equation~\eqref{eq_difficulty}. We have chosen this measure in favor of a more generic image difficulty estimation approach proposed by~\cite{Ionescu-CVPR-2016} and a domain discriminator (a CNN that discriminates between source and target samples) used by~\cite{Saito-CVPR-2019}. In order to demonstrate that our difficulty measure provides better results, we conduct a set of additional experiments. For the experiments, we first train the Faster R-CNN on original source images and on images translated by Cycle-GAN. The model is thus already adapted to the target domain and should provide more reliable labels on real target images. We next apply the model on target domain images and we divide the images into $k=3$ batches, in increasing order of difficulty. We employ here three different difficulty measures: the one that we proposed (number of objects divided by their average area), an image difficulty predictor~\citep{Ionescu-CVPR-2016} and a domain discriminator~\citep{Saito-CVPR-2019}. 

We provide the corresponding AP and mAP scores on easy, medium and hard batches for the first three benchmarks, namely Sim10k$\rightarrow$Cityscapes, KITTI$\rightarrow$Cityscapes and PASCAL VOC 2007$\rightarrow$Clipart1k, in Table~\ref{tab_suppl_difficulty_results}. Since the experiments refer to the easy-to-hard split performed on the target set, the results reported for PASCAL VOC 2007$\rightarrow$Clipart1k are completely equivalent to the results for PASCAL VOC 2007+2012$\rightarrow$Clipart1k. 
We note that, for our difficulty measure, the AP scores on the easy batch of images are higher than the AP scores on the medium batch. We observe the same behavior on the medium batch with respect to the hard batch. Unlike our approach, the image difficulty predictor does not seem to be well correlated to the AP scores on the \emph{car} class, i.e.~it does not produce the desired results for Sim10k$\rightarrow$Cityscapes and KITTI$\rightarrow$Cityscapes. Its highest AP score ($43.43\%$) on Sim10k$\rightarrow$Cityscapes is obtained for the medium batch, while its highest AP score ($40.36\%$) on KITTI$\rightarrow$Cityscapes is obtained for the hard batch. The domain discriminator is also not well correlated to the AP scores. It gives higher AP scores for the hard and the easy batches, respectively. The domain discriminator gives its lowest AP scores on the medium batch. On PASCAL VOC 2007$\rightarrow$Clipart1k, our image difficulty measure as well as the image difficulty predictor generate difficulty batches that are well correlated to the mAP scores of the Faster R-CNN. Nonetheless, our image difficulty measure gives higher differences in terms of mAP between the easy, medium and hard batches. We conclude that our difficulty measure is the only one that correlates to the AP and mAP scores on the easy, medium and hard batches, for all three benchmarks. In other words, the AP and mAP scores on the easy, medium and hard batches (as determined by our difficulty measure) decrease gradually from the easy batch to the medium batch and from the medium batch to the hard batch, respectively.

\begin{table*}[!t]
\caption{AP and mAP scores (in $\%$) of the Cycle-GAN domain adaptation framework (ii) versus our full framework (iii), both based on Faster R-CNN. Results are reported for all 20 object classes in PASCAL VOC 2007$\rightarrow$Clipart1k. Results for the baseline Faster R-CNN (i), trained only on source data, and the target domain Faster R-CNN (iv), trained on target data with ground-truth labels, are also included for comparison. The best results (except for the in-domain Faster R-CNN) are highlighted in bold.\label{tab_pvoc_07}}
\setlength\tabcolsep{0.75pt}
\begin{center}
\begin{tabular}{|c|c|c|c|c|c|c|c|c|c|c|c|c|c|c|c|c|c|c|c|c|c|}
\hline
\rotatebox{90}{Model} & \rotatebox{90}{mAP}       & \rotatebox{90}{airplane}      & \rotatebox{90}{bicycle}  & \rotatebox{90}{bird}      & \rotatebox{90}{boat}      & \rotatebox{90}{bottle}    & \rotatebox{90}{bus}       & \rotatebox{90}{car}  & \rotatebox{90}{cat}       & \rotatebox{90}{chair}     & \rotatebox{90}{cow}       & \rotatebox{90}{dining table}     & \rotatebox{90}{dog}       & \rotatebox{90}{horse}     & \rotatebox{90}{motorbike} & \rotatebox{90}{person}    & \rotatebox{90}{potted plant }  & \rotatebox{90}{sheep} & \rotatebox{90}{sofa}      & \rotatebox{90}{train}     & \rotatebox{90}{TV monitor}\\
\hline
\hline
(i)               & $18.73$   & $10.34$   & $47.44$   & $9.09$    & $12.43$   & $15.15$   & $29.14$   & $24.36$   & $9.09$   & $27.71$    & $0.00$    & $\mathbf{24.62}$  & $5.58$    & $\mathbf{27.27}$   & $41.92$ & $30.87$   & $23.15$    & $0.00$   & $1.11$   & $18.45$    & $16.78$ \\
\hline
(ii) & $21.31$   & $15.54$   & $52.76$   & $13.31$   & $12.64$   & $19.20$   & $25.12$   & $29.66$   & $9.09$   & $30.13$    & $8.59$    & $22.51$   & $9.09$    & $21.82$   & $46.56$ & $36.29$   & $20.28$    & $0.00$   & $12.17$   & $23.20$   & $18.33$ \\
(iii) & $\mathbf{27.64}$ & $\mathbf{22.25}$ & $\mathbf{61.51}$ & $\mathbf{17.85}$ & $\mathbf{16.02}$ & $\mathbf{34.76}$ & $\mathbf{34.92}$ & $\mathbf{31.97}$     & $\mathbf{9.83}$ & $\mathbf{31.54}$ & $\mathbf{26.67}$ & $23.96$   & $\mathbf{10.83}$   & $23.48$   & $\mathbf{49.79}$ & $\mathbf{55.27}$ & $\mathbf{27.27}$ & $\mathbf{5.73}$  & $\mathbf{22.10}$   & $\mathbf{25.34}$   & $\mathbf{21.62}$\\
\hline
(iv) & $33.89$   & $19.08$         & $47.27$         & $37.75$         & $25.97$         & $29.00$         & $23.36$         & $43.70$    & $19.70$         & $42.70$         & $38.93$         & $26.12$         & $21.50$         & $29.95$         & $41.16$ & $66.07$         & $33.00$         & $44.91$         & $8.50$         & $35.56$         & $43.64$\\
\hline
\end{tabular}
\end{center}
\end{table*}

\begin{table*}[!t]
\caption{AP and mAP scores (in $\%$) of the Cycle-GAN domain adaptation framework (ii) versus our full framework (iii), both based on Faster R-CNN. Results are reported for all 20 object classes in PASCAL VOC 2007+2012$\rightarrow$Clipart1k. Results for the baseline Faster R-CNN (i), trained only on source data, and the target domain Faster R-CNN (iv), trained on target data with ground-truth labels, are also included for comparison. The best results (except for the in-domain Faster R-CNN) are highlighted in bold.\label{tab_pvoc_07_12}}
\setlength\tabcolsep{0.75pt}
\begin{center}
\begin{tabular}{|c|c|c|c|c|c|c|c|c|c|c|c|c|c|c|c|c|c|c|c|c|c|}
\hline
\rotatebox{90}{Model} & \rotatebox{90}{mAP}       & \rotatebox{90}{airplane}      & \rotatebox{90}{bicycle}  & \rotatebox{90}{bird}      & \rotatebox{90}{boat}      & \rotatebox{90}{bottle}    & \rotatebox{90}{bus}       & \rotatebox{90}{car}  & \rotatebox{90}{cat}       & \rotatebox{90}{chair}     & \rotatebox{90}{cow}       & \rotatebox{90}{dining table}     & \rotatebox{90}{dog}       & \rotatebox{90}{horse}     & \rotatebox{90}{motorbike} & \rotatebox{90}{person}    & \rotatebox{90}{potted plant }  & \rotatebox{90}{sheep} & \rotatebox{90}{sofa}      & \rotatebox{90}{train}     & \rotatebox{90}{TV monitor}\\
\hline
\hline
(i)               & $26.14$   & $27.27$   & $41.70$   & $22.52$    & $20.62$   & $37.16$   & $27.73$   & $24.06$   & $15.58$   & $34.04$    & $19.51$    & $\mathbf{21.71}$  & $\mathbf{14.88}$    & $22.55$   & $39.61$ & $30.13$   & $37.77$    & $4.55$   & $17.82$   & $24.66$    & $38.88$ \\
\hline
(ii) & $30.43$   & $24.86$   & $55.65$   & $22.88$   & $21.07$   & $40.25$   & $\mathbf{42.39}$   & $29.09$   & $14.58$   & $41.91$    & $14.60$    & $20.40$   & $10.50$    & $22.50$   & $55.96$ & $41.53$   & $39.56$    & $\mathbf{9.09}$   & $20.46$   & $34.43$   & $46.80$ \\
(iii) & $\mathbf{37.83}$ & $\mathbf{41.79}$ & $\mathbf{71.97}$ & $\mathbf{30.81}$ & $\mathbf{30.88}$ & $\mathbf{51.00}$ & $40.41$ & $\mathbf{39.53}$     & $\mathbf{18.47}$ & $\mathbf{48.04}$ & $\mathbf{29.63}$ & $16.41$   & $8.15$   & $\mathbf{26.94}$   & $\mathbf{75.39}$ & $\mathbf{55.66}$ & $\mathbf{41.58}$ & $7.68$  & $\mathbf{30.58}$   & $\mathbf{38.09}$   & $\mathbf{53.60}$\\
\hline
(iv) & $33.89$   & $19.08$         & $47.27$         & $37.75$         & $25.97$         & $29.00$         & $23.36$         & $43.70$    & $19.70$         & $42.70$         & $38.93$         & $26.12$         & $21.50$         & $29.95$         & $41.16$ & $66.07$         & $33.00$         & $44.91$         & $8.50$         & $35.56$         & $43.64$\\
\hline
\end{tabular}
\end{center}
\end{table*}

The correlations between the AP/mAP scores and the difficulty levels of the easy, medium and hard batches determined by the three different difficulty measures are good indicators for knowing if these measures would be useful in a curriculum learning setting. Nevertheless, we take one step further and put these difficulty measures to the test in the context of cross-domain object detection. Each difficulty measure results in a specific curriculum learning approach. In Table~\ref{tab_suppl_diff_ablation}, we compare the standard self-paced learning approach with the curriculum self-paced learning approaches resulting from the three difficulty measures. We note that the curriculum learning strategy based on the image difficulty predictor~\citep{Ionescu-CVPR-2016} attains better results than the standard self-paced learning approach based on random sample selection. However, its improvements are lower than those of our curriculum strategy based on the number of detected objects divided by their average bounding box size. We believe that this difference appears because our method is focused precisely on the detected objects, while the image difficulty predictor is a generic approach that looks at the entire scene. Using the domain discriminator seems to be a poor decision. Indeed, the AP and mAP scores of the curriculum learning strategy based on the domain discriminator on all three benchmarks are worse than the standard self-paced learning approach. It seems that the target samples which are closer to the source samples are not necessarily those for which the object detector (trained on source data) provides the best predictions. This seems to hurt the self-paced learning process. We thus conclude that our curriculum learning strategy is the best among the three curriculum learning strategies evaluated in the experiments summarized in Table~\ref{tab_suppl_diff_ablation}.

\subsection{Results on Various Object Categories}

In order to compare with most of the related works~\citep{Chen-CVPR-2018,Khodabandeh-ICCV-2019,Saito-CVPR-2019,Shan-NC-2019,Zheng-CVPR-2020,Zhu-CVPR-2019}, we presented results on Sim10k$\rightarrow$Cityscapes and KITTI$\rightarrow$Cityscapes. Although these cross-domain benchmarks are commonly-used in literature, we and others before us reported the AP scores on the \emph{car} class only, this being the only common class between the source and the target data sets. In this context, one may naturally assume that the cross-domain object detection approach works only for a single class. In order to demonstrate that our approach works for multiple classes, we considered two additional settings, in which the source data set is either PASCAL VOC 2007 or PASCAL VOC 2007+2012 and the target data set is Clipart1k. These data sets have 20 classes in common. The mAP scores are already reported in Table~\ref{tab_results}, but some aspects still remained unanswered, e.g.~how does our model perform on each object class? In order to clarify such aspects, we report the complete results on the PASCAL VOC 2007$\rightarrow$Clipart1k benchmark in Table~\ref{tab_pvoc_07} and on the PASCAL VOC 2007+2012$\rightarrow$Clipart1k benchmark in Table~\ref{tab_pvoc_07_12}. The tables include the AP scores on all 20 classes, as well as the overall mean AP (mAP) score of each model.

The baseline Faster R-CNN (trained only on PASCAL VOC 2007) attains a mAP score of $18.73\%$. While adding source images translated by Cycle-GAN into the training is useful (the mAP improvement is around $2.58\%$), our complete approach based on Cycle-GAN and curriculum self-paced learning attains a larger improvement ($8.91\%$) in terms of mAP. Looking at the individual object classes, we note that our domain adaption approach brings performance improvements for 18 categories. The exceptions are the \emph{dining table} and the \emph{horse} classes. Compared with the Cycle-GAN domain adaptation model, we obtain better AP scores for each and every class. Remarkably, we obtain improvements higher than $24\%$ with respect to the baseline Faster R-CNN for two object classes, \emph{cow} and \emph{person}. For another four categories, \emph{airplane}, \emph{bike}, \emph{bottle} and \emph{sofa}, our improvements are higher than $12\%$. We thus conclude that the reported results demonstrate that our approach performs well on domain adaptation cases with multiple object classes, being able to improve performance over the baseline Faster R-CNN for almost all object classes (18 out of 20).

We note that Table~\ref{tab_pvoc_07} also includes the results of an in-domain Faster R-CNN trained on images from the target domain, i.e. from Clipart1k, on the last row. Perhaps surprisingly, our unsupervised domain adaptation model surpasses the in-domain Faster R-CNN on 6 object classes. This can be explained by the fact that the number of training images from the target domain is much lower (only 500). Our unsupervised domain adaptation model benefits from a much larger source data set, PASCAL VOC 2007, which contains thousands of images. These results set a good example that shows the benefit of domain adaptation methods when labeled training data from the target domain is scarce.

In Table~\ref{tab_pvoc_07_12}, we present the AP scores on each category for the models trained on PASCAL VOC 2007+2012 as source domain. Our  domain adaption method provides the highest gains for 16 out of 20 object categories. For the \emph{bus} and the \emph{sheep} classes, the Cycle-GAN domain adaptation model is better. Notably, for 8 object classes, the improvements provided by our method are $10\%$ higher than the improvements brought by the domain adaptation based on Cycle-GAN. However, there are two classes, \emph{dining table} and \emph{dog}, for which domain adaption does not seem to work at all. On the positive side, our domain adaptation approach surpasses the in-domain Faster R-CNN on 12 object classes. We therefore conclude that the results presented in Table~\ref{tab_pvoc_07_12} are consistent with those presented in Table~\ref{tab_pvoc_07}.

\section{Conclusion}
\label{sec_conclusion}

In this paper, we presented a domain adaptation method for cross-domain object detection. Our method is based on two adaptation stages. First of all, images translated from the source domain to the target domain using Cycle-GAN are added into the training set. Then, a curriculum self-paced learning approach is employed to further adapt the object detector using real target images annotated with pseudo-labels. We compared our method with several state-of-the-art-methods~\citep{Chen-CVPR-2018,Inoue-CVPR-2018,Khodabandeh-ICCV-2019,Saito-CVPR-2019,Shan-NC-2019,Zheng-CVPR-2020,Zhu-CVPR-2019} and we obtained higher absolute performance gains with respect to the corresponding Faster R-CNN baselines. Although we attained superior improvements than those reported in the recent literature~\citep{Chen-CVPR-2018,Inoue-CVPR-2018,Khodabandeh-ICCV-2019,Saito-CVPR-2019,Shan-NC-2019,Zheng-CVPR-2020,Zhu-CVPR-2019}, we notice that, in three out of four benchmarks, there are still significant performance gaps with respect to the Faster R-CNN models that are trained on the target domain with ground-truth labels (see Table~\ref{tab_results}). We believe that the performance gap can be further reduced by unifying multiple source domains during training. Data augmentation after pseudo-labeling could also play an important role in gaining additional performance. We aim to explore these directions in future work.


\section*{Acknowledgements}

This work was supported by a grant of the Romanian Ministry of Education and Research, CNCS - UEFISCDI, project number PN-III-P1-1.1-TE-2019-0235, within PNCDI III. This article has also benefited from the support of the Romanian Young Academy, which is funded by Stiftung Mercator and the Alexander von Humboldt Foundation for the period 2020-2022.

\bibliographystyle{model2-names}
\bibliography{refs}


\end{document}